\documentclass[letterpaper]{article} 
\usepackage{aaai25}  
\usepackage{times}  
\usepackage{helvet}  
\usepackage{courier}  
\usepackage[hyphens]{url}  
\usepackage{graphicx} 
\usepackage{natbib}  
\usepackage{caption} 
\usepackage{algorithm}
\usepackage{algorithmic}
\usepackage{bm}
\usepackage{amssymb}
\usepackage{amsmath}
\usepackage{color}
\usepackage{booktabs}

\newcommand{\mathvec}[1]{\mathbf{#1}}

\newcommand{\mathimg}[1]{\mathbf{#1}}
\newcommand*\samethanks[1][\value{footnote}]{\footnotemark[#1]}

\newcommand{\figref}[1]{Figure~\ref{#1}}
\newcommand{\tabref}[1]{Table~\ref{#1}}

\newcommand{\equref}[1]{Equ.~(\ref{#1})}
\newcommand{\tbdline}{\noindent\textcolor{red}{TBD TBD TBD TBD TBD TBD TBD TBD TBD TBD TBD.}\\}
\newcommand{\tbd}[1][1]{ \newcount\tmp \tmp=0 \loop \advance\tmp by 1 \tbdline \ifnum\tmp<#1 \repeat }

\usepackage{newfloat}
\usepackage{listings}
\DeclareCaptionStyle{ruled}{labelfont=normalfont,labelsep=colon,strut=off} 
\floatstyle{ruled}
\newfloat{listing}{tb}{lst}{}
\floatname{listing}{Listing}
\title{A Diffusion-Based Framework for Occluded Object Movement}
\author{
    Zheng-Peng Duan\textsuperscript{\rm 1, \rm 2}, 
    Jiawei Zhang\textsuperscript{\rm 2},
    Siyu Liu\textsuperscript{\rm 1},
    Zheng Lin\textsuperscript{\rm 5}\thanks{Corresponding authors.}, \\
    Chun-Le Guo\textsuperscript{\rm 1, \rm 3},
    Dongqing Zou\textsuperscript{\rm 2, \rm 4},
    Jimmy Ren\textsuperscript{\rm 2},
    Chongyi Li\textsuperscript{\rm 1, \rm 3}\samethanks
}
\affiliations{
    \textsuperscript{\rm 1}VCIP, CS, Nankai University \\
    \textsuperscript{\rm 2}SenseTime Research \\
    \textsuperscript{\rm 3}NKIARI, Shenzhen Futian \\
    \textsuperscript{\rm 4}PBVR \\
    \textsuperscript{\rm 5}BNRist, Department of Computer Science and Technology, Tsinghua University \\
    \{adamduan0211, zhjw1988\}@gmail.com,
    liusiyu29@mail.nankai.edu.cn,
    frazer.linzheng@gmail.com, \\
    guochunle@nankai.edu.cn, dqzou.lhi@gmail.com, rensijie@sensetime.com, lichongyi@nankai.edu.cn

}

\usepackage{bibentry}

\begin{document}

\maketitle

\begin{abstract}
Seamlessly moving objects within a scene is a common requirement for image editing,
but it is still a challenge for existing editing methods.
Especially for real-world images, 
the occlusion situation further increases the difficulty.
%
The main difficulty is that the occluded portion needs to be completed before movement can proceed.
To leverage the real-world knowledge embedded in the pre-trained diffusion models,
we propose a \textbf{Diff}usion-based framework specifically designed for \textbf{O}ccluded \textbf{O}bject \textbf{M}ovement, named \textbf{DiffOOM}.
The proposed DiffOOM consists of two parallel branches that perform object de-occlusion and movement simultaneously.
The de-occlusion branch utilizes a background color-fill strategy and a continuously updated object mask to focus the diffusion process on completing the obscured portion of the target object.
Concurrently, the movement branch employs latent optimization to place the completed object in the target location
and adopts local text-conditioned guidance to integrate the object into new surroundings appropriately.
Extensive evaluations demonstrate the superior performance of our method, which is further validated by a comprehensive user study.
%
\end{abstract}

\begin{links}
    \link{Project}{https://adam-duan.github.io/projects/diffoom/}
\end{links}



\section{Introduction}
\begin{figure}[t]
    \centering
    \includegraphics[width=0.95\linewidth]{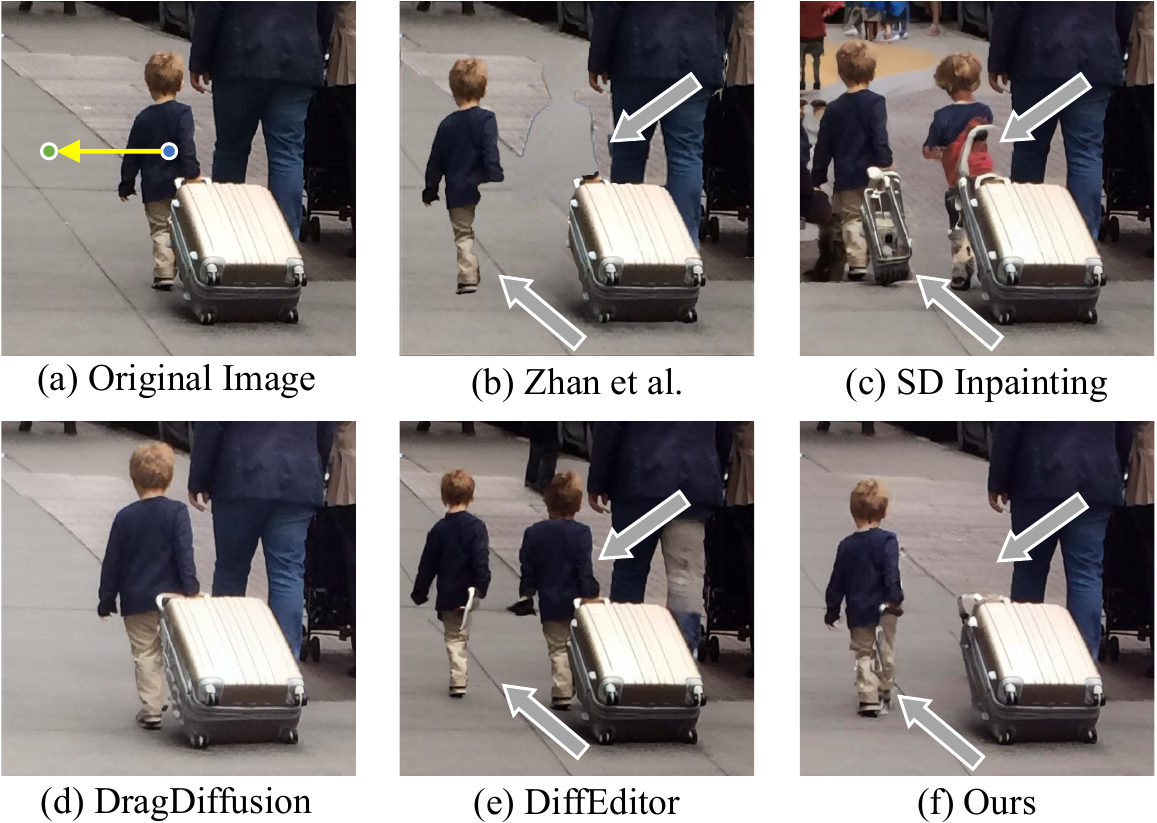}
    \vspace{-2mm}
    \caption{Comparison with other methods for occluded object movement. Given a real-world image, our method can seamlessly move the occluded object to a user-specified position while completing the occluded portion.
    }
    \vspace{-6mm}
    \label{fig:teaser}
\end{figure}

Seamlessly moving objects~\cite{avrahami2024diffuhaul} within a scene is a common requirement for image editing~\cite{nguyen2024flexedit, sajnani2024geodiffuser, epstein2023diffusion, brooks2023instructpix2pix}.
To move occluded objects,
it involves three sub-tasks: completing the obscured object, moving the object to the target position, and inpainting the original region of the moved objects.
To solve object de-occlusion, previous work~\cite{zhan2020self}
employs two separate networks: the first predicts the complete mask of the object, and the second fills in the recovered mask with reasonable content.
%
%
However, adopting discriminative networks significantly restricts the ability to generate new content,
as illustrated in~\figref{fig:teaser}(b).
Recent advances in large-scale diffusion models,
known for their powerful generative capability,
present a new opportunity to generate the occluded portion~\cite{liu2024object, zhan2024amodal, ozguroglu2024pix2gestalt, xu2024amodal}.
%
One intuitive solution is to utilize the SD Inpainting model~\cite{rombach2022high} to complete the missing regions, which is shown in \figref{fig:teaser}(c).
%
With no constraints on the generated contents,
the inpainting model may generate undesired elements rather than reconstructing the occluded portion of the target object.
Recently, diffusion-based drag-style~\cite{pan2023drag} editing methods,
such as DragDiffusion~\cite{shi2024dragdiffusion} and DiffEditor~\cite{mou2024diffeditor}, are proposed to drag objects to target positions with pre-trained diffusion models effectively.
As illustrated in \figref{fig:teaser}(d), DragDiffusion focuses primarily on content dragging and therefore struggles to move the entire object.
Although DiffEditor successfully moves the little boy to the target location in \figref{fig:teaser}(e), the occluded parts remain incomplete as de-occlusion is not considered.
%
%

Although existing diffusion-based editing methods cannot be directly employed for this task,
the comprehensive real-world knowledge embedded in large-scale diffusion models may be useful for this task.
To this end,
we propose a diffusion-based framework specifically designed for the movement of occluded objects, called DiffOOM.
%
Our method features two parallel Stable Diffusion-based branches to handle object de-occlusion and movement.
%
%

For de-occlusion,
our motivations mainly come from two aspects.
1) Diffusion models contain rich prior knowledge about the shape of various objects,
which is crucial for identifying areas that require filling.
2) Diffusion models possess strong generative abilities to complete the occluded portion with reasonable content.
%
Based on the motivations, the proposed de-occlusion branch utilizes cross-attention as well as self-attention maps to estimate the complete mask of the object,
which is utilized to guide the object occlusion region generation during the diffusion process.
To minimize the influence of irrelevant elements in the image, 
the input of the de-occlusion branch uses a color-fill strategy, where the background region of the target object is initialized as a uniform color. 
To make the visible region of the object unchanged, a latent hold strategy is adopted by replacing the diffusion-updated latent with the one from the inversion process in the visible region during the diffusion steps.
%
%
%
Besides, 
LoRA~\cite{hu2021lora} is adopted to ensure that the new content aligns with the characteristics of the target object.
%
%
%
%
%
%

With the object mask and the completed object from the de-occlusion branch, the movement branch aims to place the target object at the target location harmoniously.
%
Specifically, latent optimization minimizes the distance between the latents of the completed object and the target region, guiding the diffusion process to generate the de-occluded object in the target region.
To ensure relocated objects blend seamlessly into their new surroundings, local text-conditioned guidance is applied to the target region.
%
Another issue is to avoid filling inadequate contents into the original location of the target object like the result of DiffEditor shown in \figref{fig:teaser}(e).
To solve this issue,
we fill the original region with noise and utilize a similar mask-guided strategy to direct the diffusion process, ensuring that it inpaints the region with information from the surrounding background.
%
%

Our \textbf{contributions} can be summarized as follows:
\begin{itemize}
    \item We utilize the rich real-world knowledge embedded in pre-trained diffusion models to identify the occlusion portion of the object as well as generate the content.
    \item We introduce a dual-branch framework where the diffusion-based de-occlusion and movement branches process concurrently.
    \item Extensive experiments and a user study demonstrate the effectiveness of our method in de-occluding diverse objects and achieving satisfactory editing results.
\end{itemize}

\section{Methodology}
%
%
In real-world scenarios with occluded objects,
our goal is to enable users to relocate these objects to specified target positions while completing the occluded portions.
%
The necessary inputs for this process include the source image, denoted as $\mathimg{I}_s$, and a mask, denoted as $\mathimg{M}_v$, which highlights the visible portion of the object. 
This mask can either be provided by the user or generated through automated segmentation methods.
Additionally, the user specifies the target position by indicating the target point $\mathvec{g}$.
In the following subsections, we first introduce the preliminaries on diffusion models,
and then outline our overall framework in detail.

\subsection{Preliminaries}
\label{sec:pre}
\subsubsection{Diffusion Models}
Our method is built upon Stable Diffusion V1.5~\cite{rombach2022high},
%
%
which improves both the training and sampling efficiency of DDPM~\cite{ho2020denoising} by applying the diffusion processes in the latent space rather than pixel space.
With pre-trained encoder $\mathcal{E}$ and decoder $\mathcal{D}$,
Stable Diffusion can efficiently obtain the latent space representation $\mathimg{Z}$ of $\mathimg{X}$ by $\mathimg{Z} = \mathcal{E}(\mathimg{X})$,
and transform the latent space samples to the pixel space through $\mathcal{D}$.
%
%
To control the synthesis process through the text condition $\mathvec{c}$,
Stable Diffusion adopts the conditional denoising model $\bm{\epsilon}_\theta(\mathimg{Z}_t,t, \mathvec{c})$, where  $\mathimg{Z}_t$ is the noisy latent at timestep $t$.

\begin{figure*}[t]
    \centering
    \includegraphics[width=0.98\linewidth]{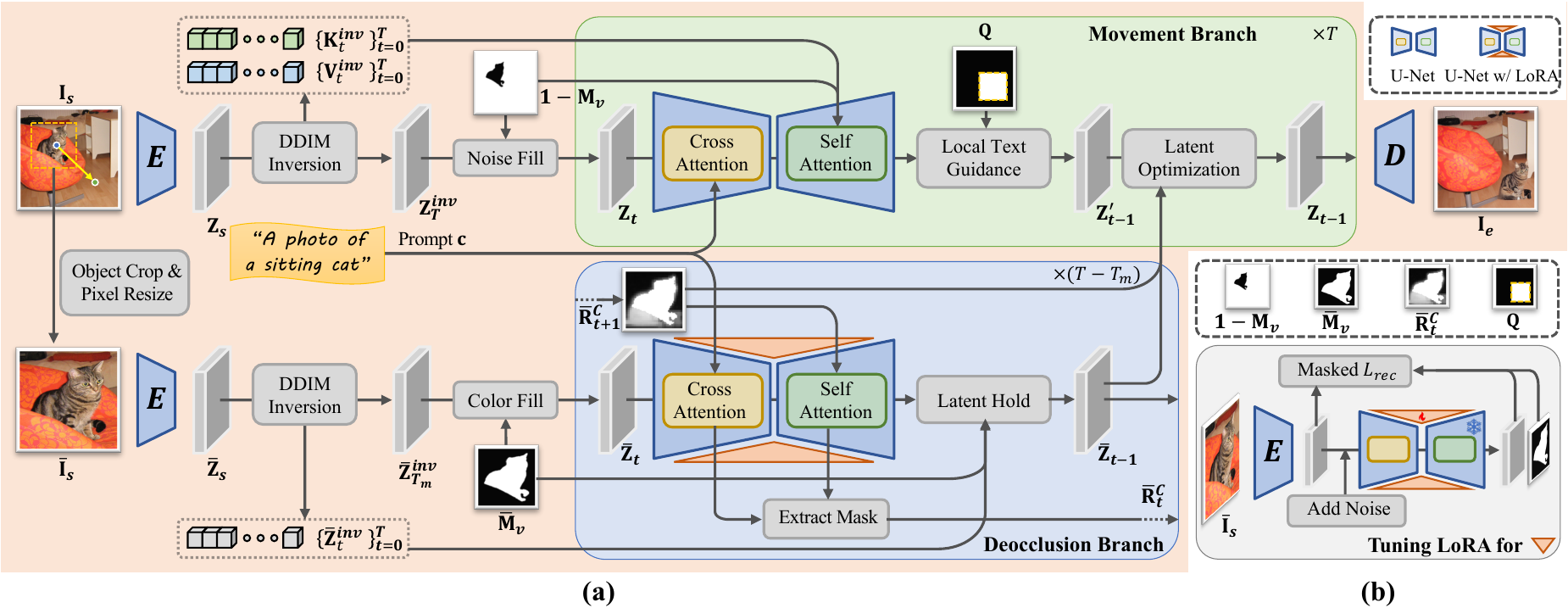}
    \vspace{-2mm}
    \caption{Overview of proposed framework (a) and LoRA tuning process (b). (a) We decouple the task of occluded object movement into de-occlusion and movement, handled by parallel branches. Both branches are built upon Stable Diffusion V1.5 and operate simultaneously. The de-occlusion branch leverages the prior knowledge within the diffusion models to complete the occluded portion, while the movement branch mainly places the completed object at the target position. (b) To ensure the content generated by the de-occlusion branch aligns with the characteristics of the target object, we equip this branch with LoRA, which is fine-tuned using a masked diffusion loss that applies exclusively to the visible portions of the object. 
    }
    \vspace{-5mm}
    \label{fig:pipeline}
\end{figure*}

\subsubsection{Attention Mechanism}
The underlying backbone of the denoising model $\bm{\epsilon}_\theta$ is a time-conditional U-Net,
which consists of a series of basic blocks.
Each basic block is equipped with a residual block, a self-attention module, and a cross-attention module sequentially~\cite{dosovitskiy2020image, vaswani2017attention}.
%
There is also a text encoder $\tau_\theta$ to project text prompt $\mathvec{c}$ of length $N$ to an intermediate representation $\tau_\theta (\mathvec{c})$.
%
At timestep $t$,
the residual block first takes the features from $(l-1)$-th basic block as input,
and generates the intermediate features $\mathimg{F}_{l,t}$.
Then, the self-attention module mines the relationship between the features and themselves,
while the cross-attention module captures the connection between visual and textual information~\cite{hertz2022prompt, tumanyan2023plug, chefer2023attend}.
%
%
Specifically,
the cross-attention map $\mathimg{A}_{l,t}^{C}$ and self-attention map $\mathimg{A}_{l,t}^{S} $ at $l$-th layer and $t$-th timestep can be obtained by
\begin{equation}
    \label{equ:attn}
    \mathimg{A}_{l,t}^{C} =\text{softmax}(\frac{\mathimg{Q}_{\mathimg{F}}\mathimg{K}^T_{\mathvec{c}}}{\sqrt{d}}) ,
    \mathimg{A}_{l,t}^{S} =\text{softmax}(\frac{\mathimg{Q}_{\mathimg{F}}\mathimg{K}^T_{\mathimg{F}}}{\sqrt{d}}),
\end{equation}
where $d$ is the dimension of features.
$\mathimg{Q}_{\mathimg{F}}$ and $\mathimg{K}_{\mathimg{F}}$ are different projections of the flattened representation of $\mathimg{F}_{l,t}$,
while $\mathimg{K}_{\mathvec{c}}$ is the projection of the text embedding $\tau_\theta (\mathvec{c})$.

\subsubsection{Refined Cross-attention Map}
As detailed in \equref{equ:attn},
the cross-attention map $\mathimg{A}_{l,t}^{C}$ illustrates the activation degree of each pixel for each text token,
while the self-attention map $\mathimg{A}_{l,t}^{S}$ captures the correlations between each pixel and others.
%
The cross-attention map related to the token representing the target object provides a rough indication of the object's location and shape,
%
%
%
and it can be further refined by utilizing the self-attention map to propagate the activated pixels to highly similar positions~\cite{nguyen2024dataset}.
%
Concretely,
we start by extracting the cross-attention map corresponding to the target object, denoted as $\Tilde{\mathimg{A}}^C_{l,t}$.
Next, we average both the cross-attention and self-attention maps at a resolution of $32\times32$ across all layers,
which can be formulated as 
\begin{equation}
    \Tilde{\mathimg{A}}_{t}^{C} =\frac{1}{L} \sum_{l=0}^{L} \Tilde{\mathimg{A}}_{l,t}^{C},
    \mathimg{A}_{t}^{S} =\frac{1}{L} \sum_{l=0}^{L} \mathimg{A}_{l,t}^{S}.
\end{equation}
We then refine the cross-attention map via:
\begin{equation}
\label{equ:map}
    \mathimg{R}_{t}^{C} = (\mathimg{A}_{t}^{S})^\lambda \Tilde{\mathimg{A}}_{t}^{C},
\end{equation}
where $\lambda$ is used to modify the influence of the self-attention map on the cross-attention map.
%
%
In common practice,
we extract the refined cross-attention map corresponding to the target object,
which we denoted as $\mathimg{R}_{t}^{C}$.

\subsection{Framework Overview}
%
We propose a diffusion-based framework specifically designed for occluded object movement. 
Our method effectively decouples this task into two sub-tasks: de-occlusion and movement, which are handled by parallel branches, as depicted in \figref{fig:pipeline}.
%
%
%
%
In the following two subsections,
we will detail the key designs of the two branches.

\subsection{Deocclusion Branch}
\label{sec:deocc}
\subsubsection{Input Preparation}
To eliminate the influence of irrelevant elements in the image, 
the de-occlusion branch takes the image patch that exclusively contains the target object as input.
Concretely,
with the visible mask $\mathimg{M}_v$ denoting the visible portion of an object, 
we compute a square bounding box that tightly encloses the object.
We denote the center point of this square as $\mathvec{b}$ and the side length as $\hat{r}$.
To ensure that the square box covers the complete object, 
we adjust the side length $r$ using a relax ratio $\eta$ by $r = \eta \cdot \hat{r} $.
Utilizing the center point $\mathvec{b}$ and the side length $r$,
we can crop the source image $\mathimg{I}_s$ and the visible mask $\mathimg{M}_v$ into square patches,
which we define as $\textbf{Crop}(\cdot, \mathvec{b}, r)$.
Since Stable Diffusion V1.5 is trained on the resolution of $512$,
we further resize these square patches,
which we denote as $\textbf{Resize}(\cdot, s)$, where $s$ represents the desired side length.
Thus,
the input image $\bar{\mathimg{I}}_s$ and the input mask $\bar{\mathimg{M}}_v$ of the de-occlusion branch can be obtained via
\begin{equation}
    \label{equ:crop}
    \{\bar{\mathimg{I}}_s, \bar{\mathimg{M}}_v\}\!=\!\textbf{Resize}(\textbf{Crop}( \{ \mathimg{I}_s, \mathimg{M}_v\}, \mathvec{b}, r)\!,\!\{512, 64\}).
\end{equation}
Clean latent can be obtained using the pre-trained encoder, 
represented as $\bar{\mathimg{Z}}_s = \mathcal{E}( \bar{\mathimg{I}}_s)$.
%
%
To establish starting points for the diffusion process,
DDIM inversion~\cite{song2020denoising} is employed, 
which maintains the consistency of the edited result.
Inspired by DragonDiffusion,
we store the intermediate noisy latents $\{\bar{\mathimg{Z}}_t^{inv}\}_{t=0}^{T}$ to provide precise guidance for the denoising process.
%
%
%
%

\subsubsection{Key Designs}
The de-occlusion branch aims to leverage the rich real-world knowledge embedded in pre-trained foundation models to complete the occluded object.
Our solution integrates two key motivations:
1) Diffusion models contain rich prior knowledge about the shape of various objects~\cite{zhan2024amodal}; and
2) Diffusion models have strong generative abilities to generate the occluded content.

\begin{figure}[t]
    \centering
    \includegraphics[width=0.82\linewidth]{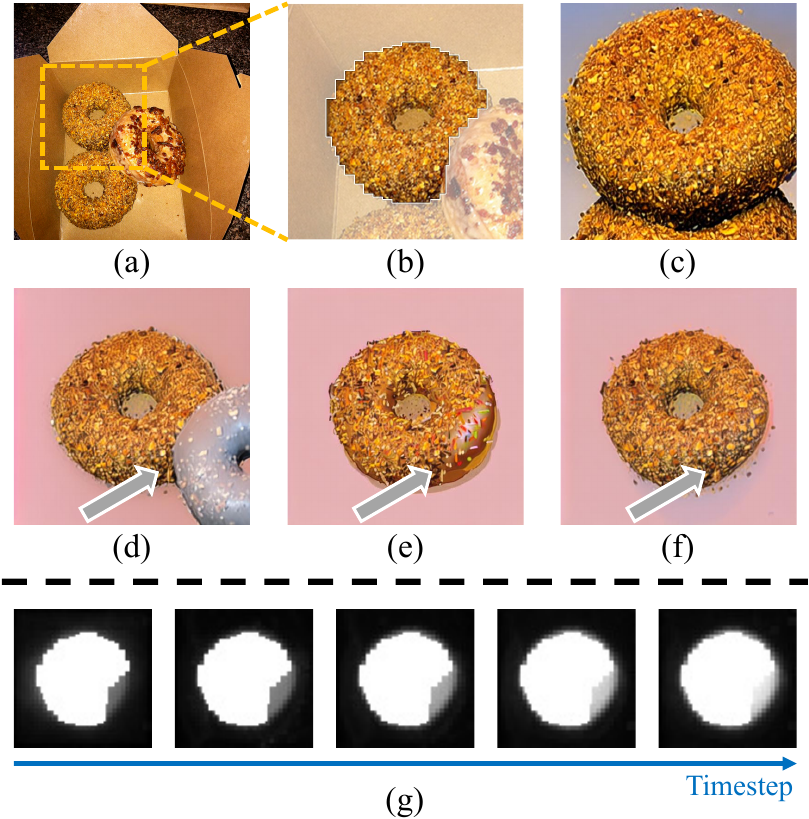}
    \vspace{-2mm}
    \caption{(a)-(b) showcase process of obtaining $\bar{\mathimg{I}}_s$ as \equref{equ:crop}.
    (b) marks $1-\bar{\mathimg{M}}_v$ with white mask. (c) - (f) are results from variants of De-occlusion Branch. (c) is generated by filling $1-\bar{\mathimg{M}}_v$ with noise as \equref{equ:noise_fill}. (d) introduces color-fill strategy as \equref{equ:color_fill}. (e) is generated under the guidance of progressively updating masks. (f) is the full Deocclusion Branch. (g) showcases the progressively updating masks based on the refined cross-attention map $\bar{\mathimg{R}}_{t}^{C}$.}
    \vspace{-4mm}
    \label{fig:deocclusion_branch}
\end{figure}

To initially validate our motivations,
we utilize the noisy latent $\bar{\mathimg{Z}}_T$ as the starting point of the diffusion process,
which is derived by filling the regions of $\bar{\mathimg{Z}}_T^{inv}$ outside the visible portion with noise.
To preserve the visible portion during the diffusion process,
we introduce a \textbf{Latent Hold} strategy by replacing the visible region $\bar{\mathimg{M}}_v$ of the intermediate sampling latent $\bar{\mathimg{Z}}'_t$ with the corresponding region from $\bar{\mathimg{Z}}_t^{inv}$ in the same time step.
This process can be formulated as 
\begin{equation}
    \label{equ:noise_fill}
    \bar{\mathimg{Z}}_t=
    \begin{cases}
    \bm{\epsilon} _T\otimes (1 - \bar{\mathimg{M}}_v) + \bar{\mathimg{Z}}_T^{inv} \otimes \bar{\mathimg{M}}_v   & \text{if}~t = T\\
    \bar{\mathimg{Z}}'_t \otimes (1 - \bar{\mathimg{M}}_v) + \bar{\mathimg{Z}}_t^{inv} \otimes \bar{\mathimg{M}}_v& \text{otherwise}
    \end{cases},
\end{equation}
where $\bm{\epsilon}_T$ is the $T$-th step noise map,
and $\otimes$ denotes the Hadamard product.
As shown in \figref{fig:deocclusion_branch}(c),
the stable diffusion process successfully generates a complete realistic donut.
%
However,
the generated donut is much larger than the original occluded donut, resulting in over-generation issues.
%

The potential reason behind over-generation is that, 
in the early stages of the diffusion, 
noise level is so high that the model fails to accurately capture the visible portion. 
Consequently, 
the model generates content freely according to the input prompt but completely ignores the visible portion.
%
%
To address this issue, we adopt two strategies:
1) We skip the early stages and start the diffusion process from the $T_m$-th step.
2) We fill the regions of $\bar{\mathimg{Z}}_{T_m}^{inv}$ outside the visible portion with uniform color.
The process can be formulated as 
\begin{equation}
    \label{equ:color_fill}
    \bar{\mathimg{Z}}_t=
    \begin{cases}
    \mathimg{J}_{T_m} \otimes (1 - \bar{\mathimg{M}}_v) + \bar{\mathimg{Z}}_t^{inv} \otimes \bar{\mathimg{M}}_v   & \text{if}~t = T_m\\
    \bar{\mathimg{Z}}'_t \otimes (1 - \bar{\mathimg{M}}_v) + \bar{\mathimg{Z}}_t^{inv} \otimes \bar{\mathimg{M}}_v& \text{otherwise}
    \end{cases},
\end{equation}
where $\mathimg{J}_{T_m}$ is a randomly colored image added with the $T_m$-th step noise.
The \textbf{Color Fill} strategy not only decreases the difficulties in capturing the visible portion, 
but also encourages the model to focus on the target object by minimizing distractions from background generation.
%

\begin{figure}[t]
    \centering
    \includegraphics[width=0.95\linewidth]{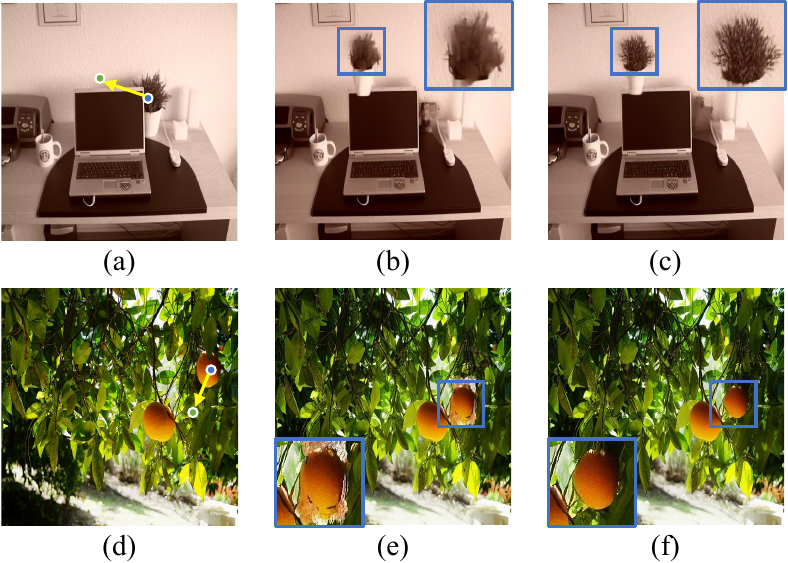}
    \vspace{-2mm}
    \caption{(a) and (d) are source images, and the others are results from variants of Movement Branch. The starting and ending points of the yellow arrows represent the original and target positions of the moved object. (b) is result with direct resizing as \equref{equ:lorg}. (c) introduces the latent resizing operation as \equref{equ:lresize}, alleviating the severe degradation. (e) and (f) are results w/o and w/ local text guidance, which helps the object integrate into surroundings more appropriately.}
    \vspace{-4mm}
    \label{fig:movement_branch}
\end{figure}

However, relying solely on these strategies does not guarantee that the model will avoid regenerating undesired elements in the obscured areas,
as shown in \figref{fig:deocclusion_branch}(d).
%
This necessitates the acquisition of a complete mask for the object to guide the diffusion process.
%
To explicitly exploit the shape priors of the diffusion model,
we extract the refined cross-attention map $\bar{\mathimg{R}}_t^{C}$ corresponding to the target object in each diffusion step.
%
%
%
The progress of $\bar{\mathimg{R}}_{t}^{C}$ throughout the diffusion process is displayed in \figref{fig:deocclusion_branch}(g).
%
It shows that the refined cross-attention map can somehow represent the complete shape of the object and it becomes more and more accurate during the diffusion process. 
To utilize this shape prior to guide the diffusion process,
we store the map $\bar{\mathimg{R}}_{t+1}^{C}$ from the previous timestep,
and send it into the self-attention module,
which restricts the attention module to query information exclusively from the target object.
%

Another challenge is the inconsistency in style between the generated and visible portions of the object, 
as depicted in \figref{fig:deocclusion_branch}(e).
Inspired by prior research~\cite{shi2024dragdiffusion, gu2024mix, avrahami2023break}, 
we conduct a style-preserving fine-tuning on the diffusion U-Net.
The finetune process is implemented with Low-Rank Adaptation (LoRA),
and the supervision is applied exclusively to the visible portion.
Equipped with LoRA, 
the diffusion model is specifically tuned to ensure the generative style aligns with the visible parts of the object.
%
%

\subsection{Movement Branch}
\label{sec:move}
\subsubsection{Input Preperation}
The input of the movement branch is the source image $\mathimg{I}_s$ and the visible mask $\mathimg{M}_v$.
After implementing the DDIM inversion on the source image $\mathimg{I}_s$,
we can obtain the output noisy latent $\mathimg{Z}_T^{inv}$, 
which sets an appropriate starting point for the movement branch to preserve the consistency between the source and edited images.
Besides, the intermediate keys $\{\mathimg{K}_t^{inv}\}_{t=0}^{T}$ and values $\{\mathimg{V}_t^{inv}\}_{t=0}^{T}$ are stored to provide guidance for subsequent diffusion process.

\subsubsection{Key Designs}
The movement branch aims to place the fully de-occluded object at the target location accurately, preserve the background information, and inpaint the original region of the moved object.
%
%
%
The input of the movement branch is initialized as $\mathimg{Z}_T^{inv} $ while the region $\mathimg{M}_v$ left at the original position of the object is filled with noise \textbf{(Noise Fill)},
which can be denoted as
\begin{equation}
    \mathimg{Z}_T= \bm{\epsilon} \otimes \mathimg{M}_v + \mathimg{Z}_T^{inv} \otimes (1-\mathimg{M}_v),
\end{equation}
where $\mathimg{Z}_T$ is the noisy latent at the $T$-th step in the movement branch. 
%
During the forward propagation of the self-attention modules in the denoising process,
we replace the keys $\mathimg{K}_t$ and values $\mathimg{V}_t$ generated from $\mathimg{Z}_t$ with the stored $\mathimg{K}_t^{inv}$ and $\mathimg{V}_t^{inv}$.
%
%
Under the guidance of $\mathimg{M}_v$,
the queries $\mathimg{Q}_t$ generated from $\mathimg{Z}_t$ are directed to retrieve the background contents from $\mathimg{K}_t^{inv}$ and $\mathimg{V}_t^{inv}$.
%
The above operations can help the background contents consistent with the input without generating undesired elements and inpaint the original region of the moved object appropriately.
%

%
%

%

%
To place the de-occluded object at the target position,
we introduce \textbf{Latent Optimization}, which strives to minimize the distance between the de-occluded object and the target region.
Through latent optimization, 
the diffusion process is guided to generate the de-occluded object in the target region.
%
%
%
Specifically, during each denoising step,
after passing the noisy latent $\mathimg{Z}_{t+1}$ through the U-Net,
we can obtain $\mathimg{Z}'_t$.
Then, we utilize the L2 distance between the complete object and the target region as the optimization objective,
which is formulated as
\begin{equation}
    \label{equ:lorg}
    \mathcal{L}_{mv} (\mathimg{Z}'_t)\!=\!\Vert \textbf{Crop}(\mathimg{Z}'_t, \mathvec{g}, r')\!-\!\textbf{Resize}(\bar{\mathimg{Z}}_t \otimes \bar{\mathimg{R}}_t^C, r') \Vert_2,
\end{equation}
where $\mathvec{g}$ is the user-specified target position.
$r'$ is the crop size in the latent space, which equals $\lceil r / 8 \rceil$ because the latent space is downsampled from the pixel space by a factor of 8.
Then we can obtain the latent $\mathimg{Z}_t$ by optimizing $\mathimg{Z}'_t$ through gradient descent, denoted as $\mathimg{Z}_t = \mathimg{Z}'_t - \gamma \frac{\partial \mathcal{L}_{mv} (\mathimg{Z}'_t)}{\partial \mathimg{Z}'_t}$.
However, 
direct resizing with bilinear interpolation is not suitable for latent space.
As depicted in \figref{fig:movement_branch} (b),
when the latent is resized in the latent space,
a significant degradation can be observed~\cite{hwang2024upsample}.
We have found a straightforward solution to this issue:
1) Decode the latent to the pixel space.
2) Perform the resizing at the pixel level using bilinear interpolation.
3) Encode the resized pixel data back into the latent space.
Thus, the latent resizing operation can be defined as 
\begin{equation}
    \label{equ:lresize}
    \textbf{L-Resize}(\mathimg{Z}, r') = \mathcal{E}(\textbf{Resize}(\mathcal{D}(\mathimg{Z}), r)).
\end{equation}
Then, we replace the resizing operation in \equref{equ:lorg} with \equref{equ:lresize}.
The result, shown in \figref{fig:movement_branch}(c), demonstrates that this approach alleviates the degradation issues.

%
As depicted in \figref{fig:movement_branch}(e),
relying solely on latent optimization is insufficient to guarantee harmonious integration.
Therefore, we leverage the priors from the diffusion model by applying classifier-free guidance at the target position.
%
We generate a mask $\mathimg{Q}$ marking the square region centered by $\mathvec{g}$ with the side length of $r$.
After passing the noisy latent $\mathimg{Z}_t$ into the denoising U-Net,
we apply text guidance~\cite{ho2022classifier} (\textbf{Local Text Guidance} in \figref{fig:pipeline}) exclusively in the target region,
which is denoted as 
\begin{equation}
    \begin{aligned}
    \label{equ:local}
    \epsilon_{pred,t}\!=\!\bm{\epsilon}_\theta(\mathimg{Z}_{t},t) \!+\!\omega \mathimg{Q} \otimes (\bm{\epsilon}_\theta(\mathimg{Z}_{t},t,\mathvec{c})\!-\!\bm{\epsilon}_\theta(\mathimg{Z}_{t},t)),
    \end{aligned}
\end{equation}
where $\omega$ represents the guidance scale.
Guided by the text prompt,
diffusion model adjusts the moved object in the target region to better align with the distribution of natural images.
This adaptation ensures that the object integrates seamlessly into new surroundings,
as observed in \figref{fig:movement_branch}(f).
%


%

\section{Experiment}
\subsection{Evaluation Dataset}

To compare the performance of our method with other existing arts,
we develop a specialized evaluation dataset derived from COCOA~\cite{zhu2017semantic} training and validation sets.
Given that our work primarily concentrates on object-level movement with occlusion,
we filter the dataset to include images that feature occluded objects of considerable size.
The final evaluation dataset comprises 120 images with a total of 150 sample objects.
For each sample, 
the visible mask $\mathimg{M}_v$ is provided by the COCO dataset~\cite{lin2014microsoft}.
%
%
As input text prompts $\mathvec{c}$ are needed for diffusion-based methods, we designed a prompt template \texttt{A photo of <category name>},
where the category name is also provided by COCO dataset.
For each sample,
we randomly set 8 different target positions,
which results in 1200 testing cases in total.

\begin{figure*}[t]
    \centering
    \includegraphics[width=0.94\linewidth]{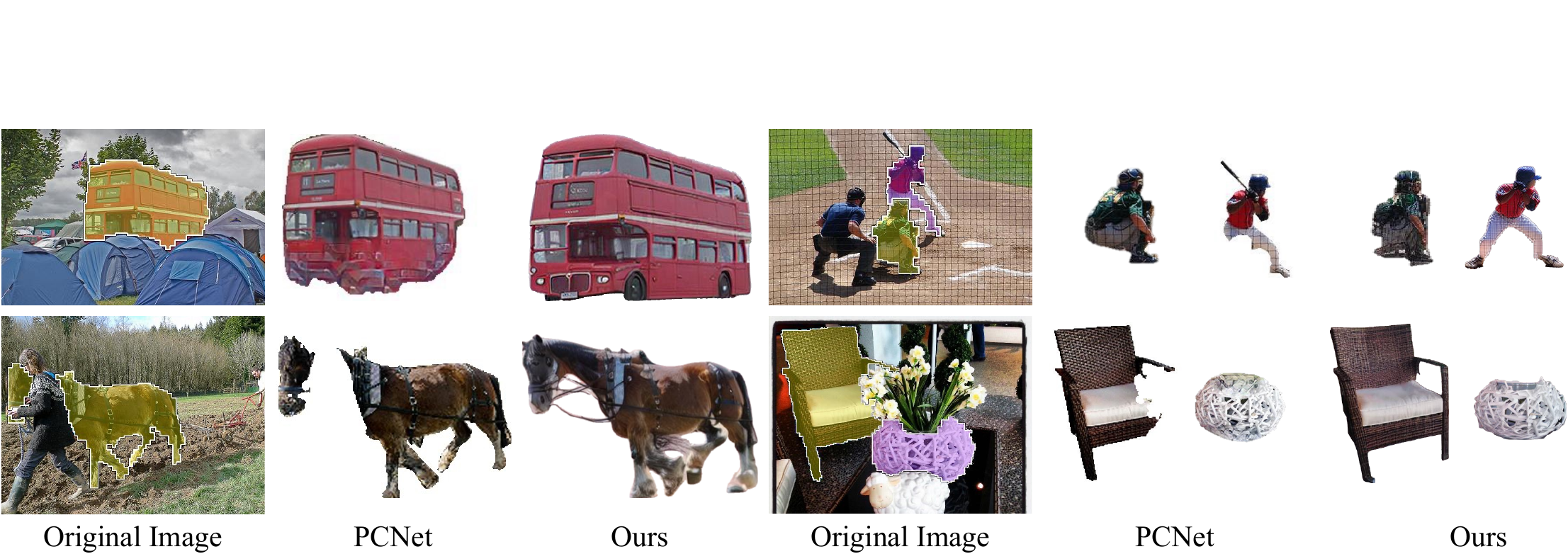}
    \caption{Qualitative comparison on de-occlusion. PCNet struggles with the completion of large-scale occlusion and complex objects, while our method can generate high-quality content consistent with the target object.
    }
    \label{fig:result_deocclusion}
    \vspace{-2mm}
\end{figure*}

\subsection{Comparison on De-occlusion}
\subsubsection{Evaluation Metrics}
For quantitative evaluation of the realism of the de-occluded objects,
we adopt the KID score~\cite{binkowski2018demystifying},
comparing the de-occluded objects with the ground-truth complete objects in the COCOA dataset.
Note that all object images are set against a white background to ensure a fair comparison.
Following DreamBooth~\cite{ruiz2023dreambooth},
we also use the CLIP-T metric to evaluate the prompt fidelity,
which is measured as the average cosine similarity between prompt and image CLIP~\cite{radford2021learning} embeddings.

\subsubsection{Comparison with Other Methods}
To validate the effectiveness of our de-occlusion branch,
we conduct a comparison with the object de-occlusion method, PCNet~\cite{zhan2020self}.
%
%
\tabref{tab:com_deocc} reports the quantitative comparison results,
demonstrating that our method outperforms PCNet in terms of image realism and prompt fidelity.
Furthermore,
the qualitative comparisons illustrated in \figref{fig:result_deocclusion} highlight that our method can produce more satisfactory results than PCNet.
Although PCNet can achieve relatively good results for simple cases, 
it struggles with the completion of large-scale occlusions and complex subjects.
Leveraging the extensive real-world knowledge embedded in pre-trained diffusion models, 
our method excels at de-occluding complex objects and generating high-quality content.

\subsection{Comparison on Occluded Object Movement }
\subsubsection{Evaluation Metrics}
For the evaluation of the occluded object movement,
we mainly focus on the original position (OP) and the target position (TP) of the moved object.
We adopt three evaluation metrics:
DINO-OP, DINO-TP, and CLIP-TP.
DINO-OP measures whether the target object leaves no residual at the original position.
To this end,
we crop the box area around the original position for both the source image $\mathimg{I}_s$ and the edited result $\mathimg{I}_e$, 
and use DINOv2~\cite{oquab2023dinov2} to measure the similarity between the box areas.
A higher DINO-OP score indicates that the target object remains at the original position, 
which is undesirable.
To measure whether the target object is indeed moved to the target position, 
we crop the box area around the original position in $\mathimg{I}_s$ and around the target position in $\mathimg{I}_e$.
Similarly,
we utilize DINOv2 to measure the cosine similarity between the crops.
Higher DINO-TP scores represent that the target object is successfully moved to the target position.
Additionally,
we also compare the similarity of the image CLIP embeddings of these two patches,
which reflects the object's fidelity and harmony with its surroundings.
%

\subsubsection{Comparison with Other Methods}
In this section,
we compare our method against feasible editing methods for occluded object movement.
We divide these methods into four categories.
1) Paint-By-Example (PBE)~\cite{yang2023paint} and AnyDoor~\cite{chen2024anydoor} are designed to add an object to an image.
To adapt them for object movement,
we apply the image inpainting at the original position of the object,
and then paste the object at the target position.
2) SD Inpainting~\cite{rombach2022high} is the state-of-the-art inpainting method,
which we can also convert into occluded object movement.
With the visible mask $\mathimg{M}_v$, we can extract the object and paste it to the target position.
By masking the original position and the areas around the target position,
the model is forced to fill these regions with reasonable content.
3) DragDiffusion~\cite{shi2024dragdiffusion} is a point-dragging editing method,
which we adapt for object movement by selecting multiple points on the target object.
4) DragonDiffusion~\cite{mou2023dragondiffusion} and DiffEditor~\cite{mou2024diffeditor} can be directly applied since it can tackle object movement.
%
\begin{table}[tb]
\centering
\footnotesize
\begin{tabular}{c|c|c}
\toprule
Method      & KID $\downarrow$   & CLIP-T $\uparrow$ \\
\midrule
PCNet & 0.0186 & 29.57  \\
Ours        & $\bm{0.0142}$ & $\bm{30.49}$  \\
\bottomrule
\end{tabular}
\caption{Quantitative comparison on de-occlusion. Our method outperforms PCNet for both image realism and prompt fidelity according to KID and CLIP-T, respectively.}
\vspace{-5mm}
\label{tab:com_deocc}
\end{table}

\begin{figure*}[t]
    \centering
    \includegraphics[width=0.94\linewidth]{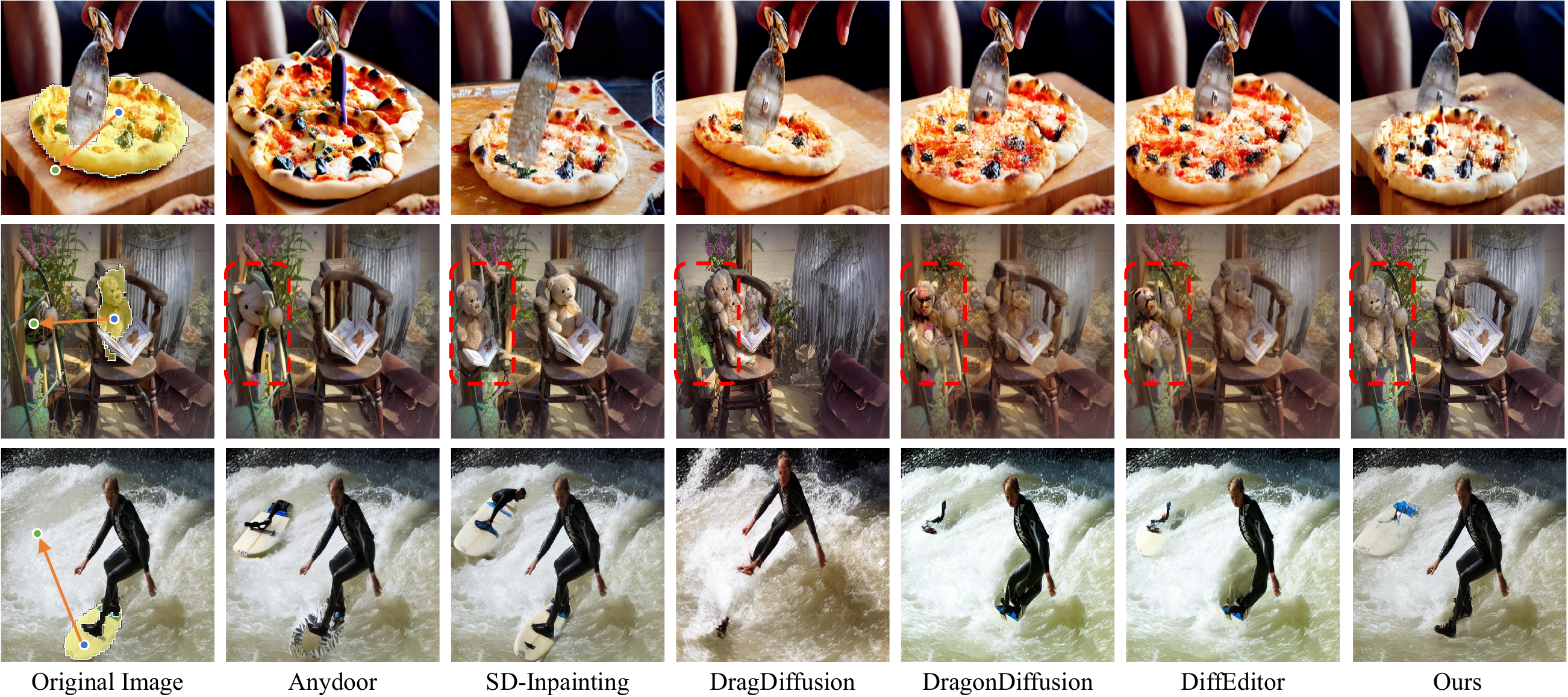}
    \caption{Qualitative comparison on occluded object movement. The target objects are marked by yellow masks. The starting and ending points of the orange arrows represent the original and target positions of the moved object.
    }
    \vspace{-3mm}
    \label{fig:result_movement}
\end{figure*}

The quantitative comparison is reported in \tabref{tab:com_mv}.
PBE and AnyDoor receive low scores in DINO-TP due to their inability to preserve the original pose and appearance of the edited object.
SD Inpainting achieves high performance on DINO-TP because it effectively relocates the object to the target position by directly pasting it.
However, this leads to a lack of harmony between the object and its new surroundings, reflected by the low CLIP-TP score.
Additionally, it may generate undesired objects at the original position, leading to a poor DINO-OP score.
DragDiffusion performs poorly across all three metrics as it primarily focuses on content dragging rather than object movement.
DragonDiffusion and DiffEditor encounter significant issues with residual artifacts, 
which negatively impacts their DINO-OP scores.
Our method outperforms the baselines,
which is also supported by the qualitative comparison in \figref{fig:result_movement}.

\begin{table}[tb]
\centering
\footnotesize
\begin{tabular}{c|c|c|c}
\toprule
Method          & DINO-OP $\downarrow$ & DINO-TP $\uparrow$ & CLIP-TP $\uparrow$\\
\midrule
PBE             & 0.575   & 0.690   & 0.950  \\
AnyDoor         & 0.651   & 0.728   & 0.952  \\
SD Inpainting   & 0.745   & $\bm{0.742}$   & 0.949  \\
DragDiff   & 0.647   & 0.706   & 0.933  \\
DragonDiff & 0.673   & 0.730   & 0.957  \\
DiffEditor      & 0.678   & 0.731   & 0.958  \\
Ours            & $\bm{0.561}$   & $\bm{0.742}$   & $\bm{0.960}$ \\
\bottomrule
\end{tabular}
\caption{Quantitative comparison on occluded object movement. Our method outperforms the compared methods.}
\label{tab:com_mv}
\vspace{-2mm}
\end{table}

\begin{table}[tb]
\centering

\footnotesize
\begin{tabular}{c|c|c|c}
\toprule
Ours vs.       & OP     & TP    & Realism \\
\midrule
AnyDoor       & 86.8\% & 84.3\% & 85.0\%  \\
SD-Inpainting & 78.0\% & 79.3\% & 75.8\%  \\
DragDiffusion & 81.0\% & 80.5\% & 80.9\%  \\
DiffEditor    & 80.8\% & 79.0\% & 78.8\%  \\
\bottomrule
\end{tabular}
\caption{User Study. Users are asked to choose a better result (Ours vs. the baseline) in terms of: 1) No residual artifacts at the original position (OP).
2) Placement of the completed object at the target position (TP).
3) Maintenance of image realism (Realism).
The numbers indicate the winning rate of our method over the compared method.}
\vspace{-5mm}
\label{tab:user_study}
\end{table}

\subsubsection{User Study}
To further evaluate the visual quality,
we invite 60 volunteers for a user study.
We select four methods (AnyDoor, SD Inpainting, DragDiffusion, and DiffEditor) to make the comparison,
and each method generated 40 edited results.
%
%
For each comparison, volunteers are required to choose whether our result is better than one of the methods.
%
As shown in \tabref{tab:user_study},
our method is preferred over these methods with a higher winning percentage.

\begin{table}[tb]
\centering
\footnotesize
\begin{tabular}{c|c|c|c}
\toprule
Method                  & DINO-OP $\downarrow$ & DINO-TP $\uparrow$ & CLIP-TP $\uparrow$\\
\midrule
Ours            & $\bm{0.561}$   & $\bm{0.742}$   & $\bm{0.960}$ \\
\midrule
w/o CF         & 0.565   & 0.717   & 0.957  \\
w/o AG           & 0.612   & 0.731   & 0.959  \\
w/o LoRA                & 0.564   & 0.733   & 0.958  \\
w/o LR       & 0.561   & 0.695   & 0.958  \\
w/o LTG & 0.563   & 0.732   & 0.955  \\
\bottomrule
\end{tabular}
\caption{Ablation Study. We conduct ablation study on the following components:
1) w/o Color Fill Strategy (CF),
2) w/o Attention Guidance (AG),
3) w/o LoRA,
4) w/o Latent Resizing(LR), and
5) w/o Local Text Guidance (LTG).
}
\label{tab:ablation}
\vspace{-4mm}
\end{table}

\subsection{Ablation Study}
We conduct ablation study for the following components and the results are reported in \tabref{tab:ablation}.
1) \textbf{Color Fill Strategy (CF)}:
Removing CF harms the DINO-TP score,
as it may cause the problem of over-generation,
which reduces the similarity to the original object.
%
2) \textbf{Attention Guidance (AG)}:
For the de-occlusion branch,
without the progressive updating mask restricting the attention module to query information exclusively from the target object,
it may generate undesired elements, resulting in a lower DINO-TP score.
For the movement branch,
removing the mask indicating the background region,
the diffusion model may re-generate the original object at the original position,
resulting in a high DINO-OP score.
3) \textbf{LoRA}:
Removing the LoRA causes the generated portion inconsistent with the original portion,
which reduces the DINO-TP score.
4) \textbf{Latent Resizing(LR)}:
When removing the LR and adopting the pixel resizing directly,
the target object encounters severe degradation,
significantly lowering the DINO-TP score.
5) \textbf{Local Text Guidance (LTG)}:
LTG aims to integrate the target object seamlessly into new surroundings.
Therefore, removing LTG harms the harmony of the target region,
which is reflected in the lower score of CLIP-TP.
%

\begin{figure}[t]
    \centering
    \includegraphics[width=0.95\linewidth]{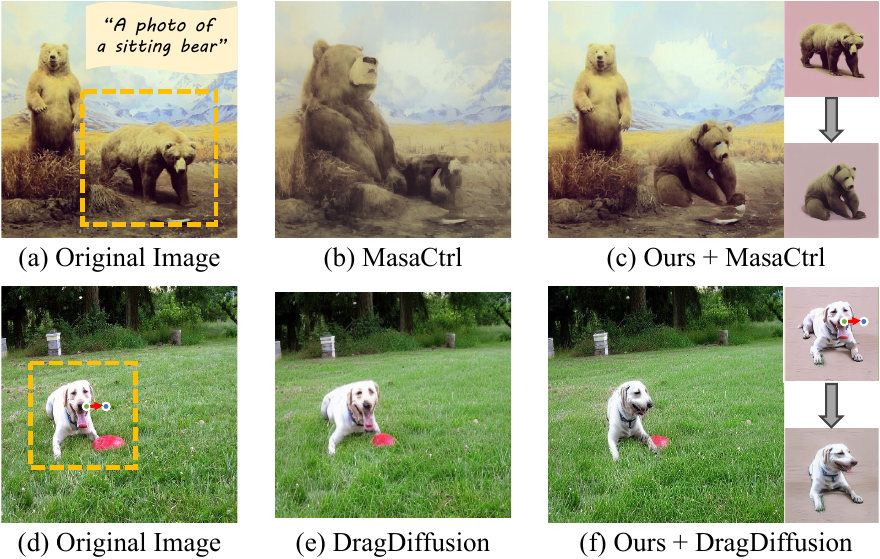}
    \caption{Examples of integrating our framework with MasaCtrl and DragDiffusion. Our framework enables two editing methods to perform precise editing in complex scenes.}
    \vspace{-7mm}
    \label{fig:masa_drag}
\end{figure}

\subsection{Integration with Other Methods}
\label{sec:inte}
%
Our dual-branch framework allows for the decomposition of the object from its background, 
enabling focused edits directly on the object.
The de-occlusion branch is compatible with most existing editing methods, enhancing their ability to perform precise edits in complex scenes.
\figref{fig:masa_drag} provides two examples where our de-occlusion branch is integrated with two typical editing methods, text-conditioned method (MasaCtrl~\cite{cao2023masactrl}) and drag-style method (DragDiffusion).
%
%
The main advantages of our framework in enhancing these methods are:
1) In scenarios with multiple similar objects, our framework allows for the editing of a specific object without affecting others.
2) By isolating the object from a complex background, 
our method permits more precise control over the object.

\section{Conclusion}
In this paper, we propose a diffusion-based framework for occluded object movement.
We demonstrate that extensive prior knowledge within diffusion models is helpful for this task.
In our dual-branch network,
the de-occlusion branch completes the occluded portion of the target object, while the movement branch places the restored object at the target position.
Moreover,
our framework can be integrated with existing editing methods, 
enabling them to perform precise editing in complex scenes.
%
%
We hope that our framework will serve as a valuable tool for image editing tasks in the future.

\section{Acknowledgments}
This work is funded by the National Natural Science Foundation of China (62306153), the Fundamental Research Funds for the Central Universities (Nankai University, 070-63243143), the China Postdoctoral Science Foundation (GZB20240357, 2024M761682), and Shui Mu Tsinghua Scholar (2024SM079).
The computational devices of this work is supported by the Supercomputing Center of Nankai University (NKSC).

\bibliography{main}

\end{document}